\documentclass[letterpaper, 10 pt, journal, twoside]{ieeetran}

\IEEEoverridecommandlockouts                              

\title{\LARGE \bf
FLARE: Agile Flights for Quadrotor Cable-Suspended Payload \\ System via Reinforcement Learning}

\usepackage{graphicx}
\usepackage{stfloats}
\usepackage{placeins}
\usepackage{subfigure}
\usepackage{amsmath}
\usepackage{mathtools}
\usepackage{float}
\usepackage{amsfonts,amssymb}
\usepackage{multirow}
\usepackage{makecell}
\usepackage{tabulary}
\usepackage{hyperref}
\usepackage{booktabs}
\usepackage{siunitx}   
\usepackage{threeparttable}
\usepackage{cuted}
\usepackage{makecell}
\hypersetup{hidelinks,
	colorlinks=true,
	allcolors=blue,
	pdfstartview=Fit,
	breaklinks=true}
\DeclareMathOperator*{\argmax}{arg\,max}
\usepackage[font=footnotesize]{caption}
\usepackage[font=scriptsize]{subcaption}

\author{Dongcheng Cao, Jin Zhou, Xian Wang, and Shuo Li
\thanks{The authors are with the College of Control Science and Engineering, Zhejiang University, Hangzhou 310027, China (e-mail: shuo.li@zju.edu.cn, cdc11@zju.edu.cn).}%
\thanks{Project page: https://bei11hai.github.io/Flare-web/.}
\thanks{Code is available at: https://github.com/BEI11HAI/Flare.}
}

\begin{document}

\maketitle

\begingroup
\setlength{\abovecaptionskip}{-1pt}
\setlength{\belowcaptionskip}{-5pt}
    \begin{strip}
        \vspace*{-3.8cm}
        \centering
        \includegraphics[width=\textwidth,trim = 0 225 238 0, clip]{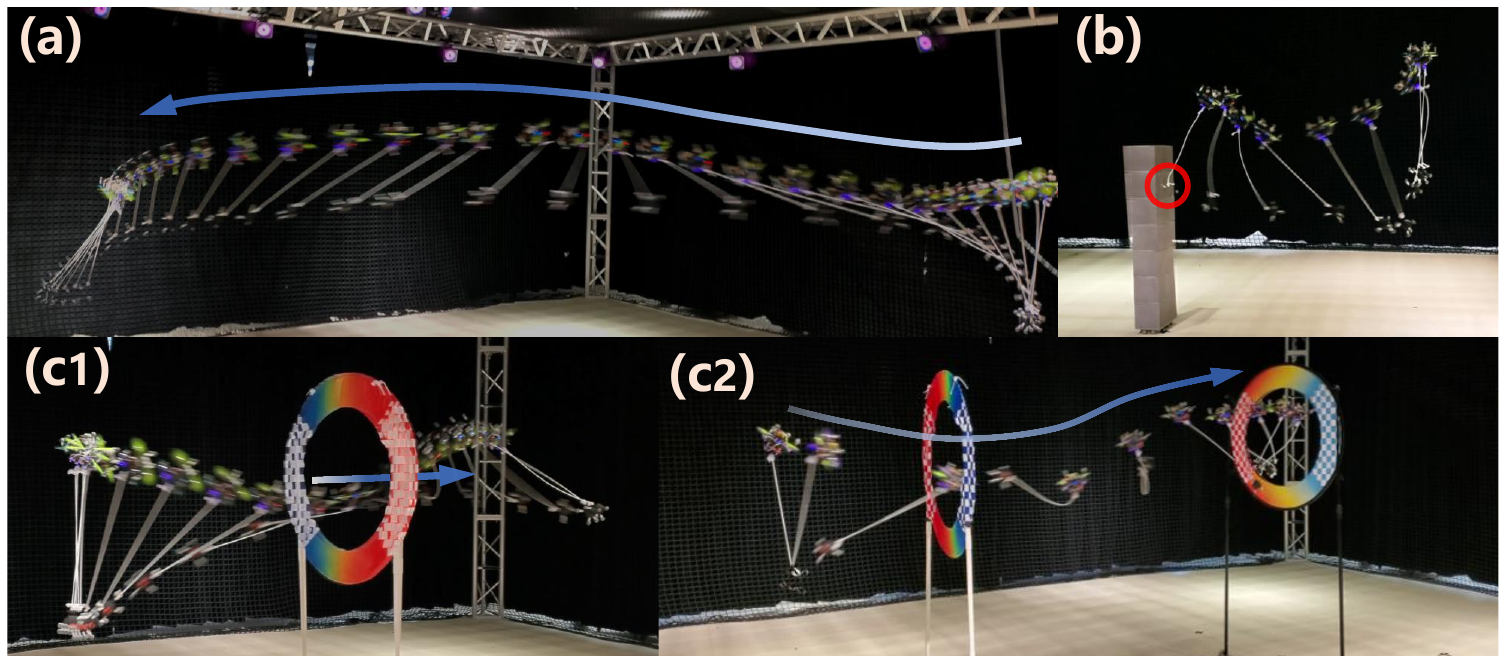}
        \vspace{-0.6em}
        \captionof{figure}{ \textbf{Autonomous Agile Flights of a Quadrotor Cable-suspended Payload System via Reinforcement Learning Policy.} The system demonstrates its capabilities in three challenging real-world scenarios: (a) agile waypoint passing, (b) payload targeting, and (c1, c2) one-time and sequential gate traversal.}
        \label{real-exp-fig}
        \vspace{-0.25em}
    \end{strip}
\endgroup

\begin{abstract}
Agile flight for the quadrotor cable-suspended payload system is a formidable challenge due to its underactuated, highly nonlinear, and hybrid dynamics. Traditional optimization-based methods often struggle with high computational costs and the complexities of cable mode transitions, limiting their real-time applicability and maneuverability exploitation. In this letter, we present FLARE, a reinforcement learning (RL) framework that directly learns an agile navigation policy from high-fidelity simulation. Our method is validated across three designed challenging scenarios, notably outperforming a state-of-the-art optimization-based approach by a 3x speedup during gate traversal maneuvers. Furthermore, the learned policies achieve successful zero-shot sim-to-real transfer, demonstrating remarkable agility and safety in real-world experiments, running in real time on an onboard computer. 
\end{abstract}

\hypersetup{hidelinks,
	colorlinks=true,
	allcolors=black,
	pdfstartview=Fit,
	breaklinks=true}

\vspace{-1em}
\section{INTRODUCTION}
Over the past few decades, agile flight technology for quadrotors has advanced rapidly. More recently, there appears to be growing research interest in agile flight of quadrotor-based aerial robotic platforms. Among these platforms,  cable-suspended payload systems have attracted considerable attention, with a burgeoning field of research exploring their promising applications \cite{palunko2012agile}, \cite{tang2015mixed}, \cite{li2021pcmpc}.

Navigating the suspended payload system has always been a challenging problem, let alone achieving agile flights. The inherent strong nonlinearity and underactuation of the system impose fundamental difficulties in planning and control. Moreover, the system exhibits hybrid dynamic characteristics, where the taut or slack mode of the cable significantly alters the system dynamics, presenting additional challenges. Driven by this concern, some early studies \cite{palunko2012trajectory}, \cite{sreenath2013trajectory} focus primarily on minimizing cable oscillations to ensure safety at the expense of flexibility and agility.

To solve this problem, various approaches based on optimization are proposed, aiming to further exploit the system's maneuverability for more agile flights. These approaches converge on the idea of planning motions by solving an optimal control problem (OCP) in a model-based manner. Within these approaches, the cable swing dynamics is actively leveraged to facilitate agile flights of the suspended payload system \cite{tang2018aggressive}, even achieving collision-free agile flights in cluttered environments \cite{son2020real}, \cite{li2023autotrans}. Despite their impressive results, these works are predicated exclusively on the taut-cable paradigm, without considering the slack mode. To address this limitation, a unified system model is introduced that incorporates complementarity constraints to describe mode transitions \cite{foehn2017fast}. Consequently, some works exploit the hybrid modes of this system to generate or track agile trajectories involving automatic mode switching \cite{zeng2020differential}, \cite{sarvaiya2024hpa}. More recently, Wang et al. achieve flights with automatic mode-switching through a systematic planning and control framework, \textcolor{black}{named Impactor \cite{wang2024impact}}, which is remarkably validated in a gate-traversal scenario. However, the corresponding high computational costs prevent real-time trajectory planning in these works \cite{foehn2017fast, zeng2020differential, wang2024impact}.

Notably, Reinforcement Learning (RL) method offers a promising alternative in a data-based manner, with extensively growing applications in quadrotor agile flights in recent years \cite{song2021autonomous}, \cite{kaufmann2023champion}, \cite{wang2025dashinggoldensnitchmultidrone}. Indeed, applying RL methods to achieve agile flights for the suspended payload system offers three principal advantages. First, the model-free nature of RL eliminates reliance on gradients of the dynamic model, thereby avoiding the significant difficulties encountered by traditional gradient-based optimization methods when handling nonsmooth problems. Second, the stochastic exploration mechanisms embedded in most RL algorithms grant them the ability to escape local minima, crucial for solving non-convex optimization problems. Third, the trained policies are typically computationally lightweight, permitting their onboard deployment for real-time planning.

Meanwhile, RL methods have been preliminarily explored for the navigation of the suspended payload system. Pioneering RL implementations prioritize the swing minimization \cite{palunko2013reinforcement}, \cite{faust2017automated}, achieving safety guarantees while compromising on agility. Another work augments the RL framework with the meta-learning technique, facilitating rapid mid-flight adaptation to unknown physical properties \cite{belkhale2021model}, achieving robust performance across varying payload weights and cable lengths. However, the application of RL methods to achieve truly agile flights of the suspended payload system still remains unexplored.

Hence, in this letter, we tackle the challenge of agile flight for the suspended payload system by formulating it within a tailored RL framework. We employ a model-free RL method in a high-fidelity simulation platform. \textcolor{black}{Crucially, the trained policy establishes a direct state-to-control mapping that inherently handles the system's hybrid dynamics (slack/taut modes), circumventing the heavy computational burden of mode-switching optimization.} The lightweight network enables high-frequency inference solely with onboard resources. Overall, the main contributions of this letter are summarized as follows:

\begin{enumerate}
    \item We propose an RL framework for the agile navigation of the quadrotor cable-suspended payload system, featuring observations and reward functions tailored \textcolor{black}{for successful zero-shot transfer}.
    \item We design three representative application scenarios and \textcolor{black}{present, to the best of our knowledge, the first real-time replanning framework that explicitly handles the system's hybrid dynamics (taut/slack modes).}
    \item We validate our proposed method \textcolor{black}{through extensive experiments along with comprehensive analysis. The results demonstrate successful zero-shot sim-to-real transfer and real-time onboard execution, notably outperforming the state-of-the-art optimization method.}
\end{enumerate}

\vspace{0.5em}
\section{METHODOLOGY}
\begin{figure*}
     \centering
    \includegraphics[width=0.95\textwidth,trim = 0 270 3 0, clip]{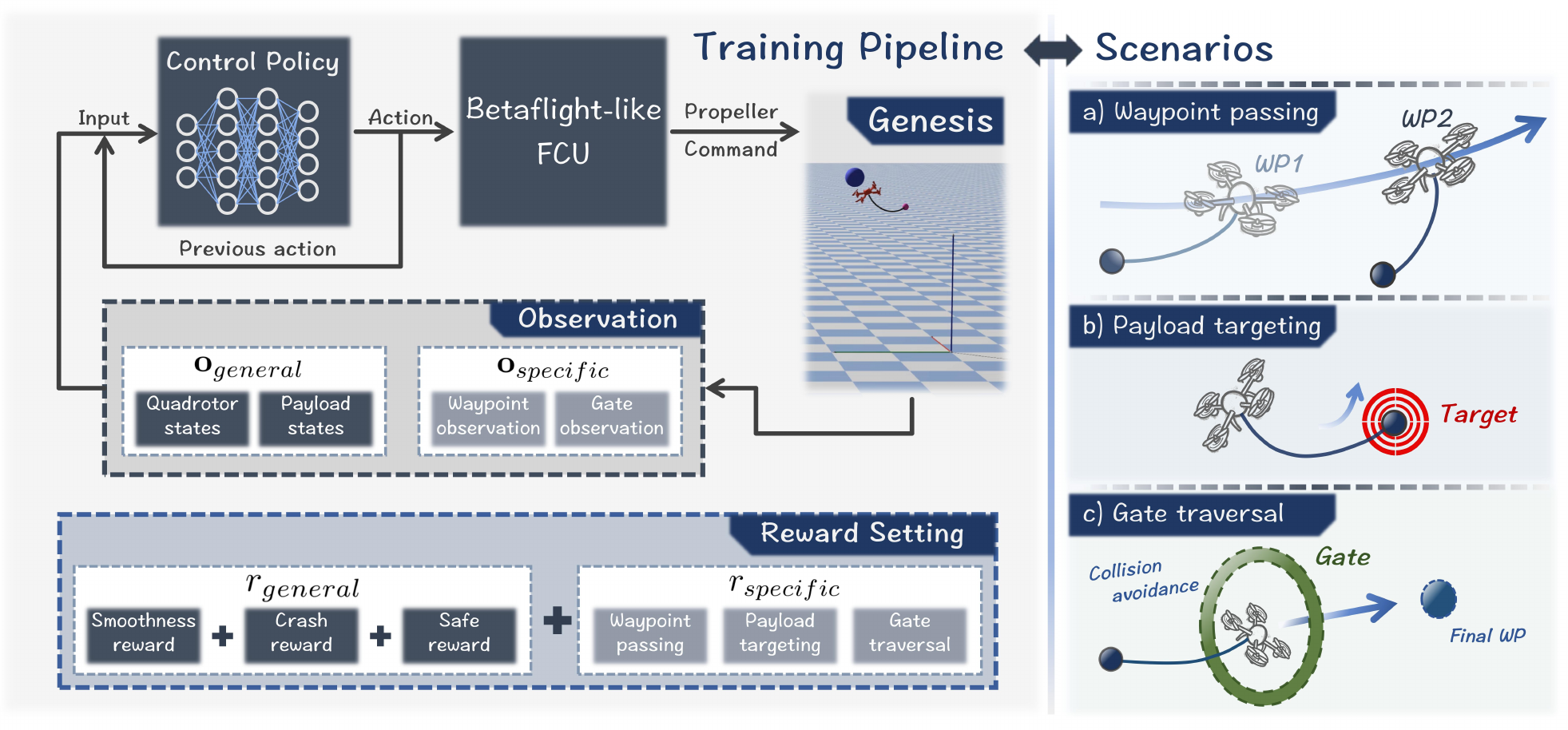}
    \vspace{-0.6em}
    \caption{  Overview of our proposed approach. We design three representative application scenarios focused on motion planning of cable-suspended payload system, each with a unique goal. In each scenario, a control policy is learned with customized observations and reward setting in Genesis simulator \cite{Genesis}.}
    \label{training pipeline}
    \vspace{-0.8em}
\end{figure*}

This section details our agile but safe motion planning approach for the quadrotor suspended payload system. We first present the system's complex hybrid dynamics, the challenges of which motivate our model-free RL approach. We then formulate the motion planning problem as a Markov Decision Process and design three representative application scenarios: agile waypoint passing, payload targeting, and gate traversal. Finally, we describe more training details, including a domain randomization strategy to enhance the policy's real-world generalization.

\subsection{Problem formulation}
We formulate the motion planning problems of the suspended payload system as a general infinite-horizon Markov Decision Process (MDP), defined as \(\mathcal{M}=(\mathcal{S}, \mathcal{A}, \mathcal{P}, \mathcal{R}, \gamma)\). Here, \(\mathcal{S}\) denotes the state space, \(\mathcal{A}\) the action space, \(\mathcal{P}\) the transition probability, \(\mathcal{R}\) the reward function, and \(\gamma\) the discount factor. Within this MDP framework, we aim to optimize the parameters \(\psi\) of a policy network \(\pi_{\psi}\) to maximize the expected discounted return:

\begin{equation}
\pi_{\psi}^{*} = \argmax_{\pi} ~\mathbb{E}\left[ \sum_{t=0}^{\infty} \gamma^t r(t) \right]
\end{equation}
where \(\gamma \in [0,1)\) is the discount factor, and \(r(t)\) represents the immediate reward at time step \(t\). Within this framework, we develop three distinct application scenarios focused on suspended system motion planning, each presenting unique challenges. Following this, we define the observations, actions, and reward functions, as the general framework illustrated in Fig. \ref{training pipeline}.

\subsubsection{Observations and Actions}
For all three scenarios, the observation vector $\mathbf{o}$ consists of two parts, defined as follows:
\begin{equation}
    \mathbf{o}=[\mathbf{o}_{\textit{general}},\mathbf{o}_{\textit{specific}}]^T,\\
\end{equation}
where $\mathbf{o}_{\textit{general}}$ represents the general observations shared across all three scenarios, while $\mathbf{o}_{\textit{specific}}$ consists of scenario-specific observations that will be detailed in section \ref{scenario customization}.

Specifically, the general observation component, $\mathbf{o}_{\textit{general}} = [\mathbf{o}_q,\mathbf{o}_p]^T$, comprises the observations of quadrotor states $\mathbf{o}_q$ and the payload states $\mathbf{o}_p$. The quadrotor state observation, $\mathbf{o}_q = [\mathbf{v}_q, \operatorname{vec}{\left(\mathbf{R}(\mathbf{q})\right)}] \in \mathbb{R}^{12}$, includes quadrotor velocity $\mathbf{v}_q$ and flattened rotation matrix $\operatorname{vec}{\left(\mathbf{R}(\mathbf{q})\right)}$ derived from quadrotor quaternion $\mathbf{q}$.  The payload observation $\mathbf{o}_p = [\phi, \theta]^\top \in \mathbb{R}^2$ encodes its deviation angles from the negative z-axis $-\mathbf{z}_{B}$ of the quadrotor body frame $\mathcal{B}$, (Fig. \ref{cable-angle}), where:
\begin{equation}
\begin{aligned}
\phi &= \mathrm{atan2}(y_b, -z_b), ~~\theta = \mathrm{atan2}(x_b, -z_b), 
\end{aligned}
\end{equation}
with $[x_p^b,y_p^b,z_p^b]^T$ being the positional coordinates of the payload in body frame $\mathcal{B}$. These components of the general observation $\mathbf{o}_{\textit{general}}$ provide the policy with essential information about the system's current states.

The policy network $\pi_\psi$ generates normalized actions $\mathbf{a} = [\tilde{T}_{\text{cmd}}, \tilde{\boldsymbol{\omega}}_{\text{cmd}}]$, \textcolor{black}{corresponding to the normalized collective thrust and body rates, respectively. These are transformed to physical control inputs for the quadrotor,} specifically the thrust $T_{\text{cmd}}$ and body rates $\boldsymbol{\omega}_{\text{cmd}}$ through: 
\begin{equation}
T_{\text{cmd}} = \frac{\tilde{T}_{\text{cmd}} + 1}{2}T_{\text{max}}, \quad
\boldsymbol{\omega}_{\text{cmd}} = \tilde{\boldsymbol{\omega}}_{\text{cmd}} \boldsymbol{\omega}_{\text{max}},
\end{equation}
where $T_{\text{max}}$ and $\boldsymbol{\omega}_{\text{max}}$ represent the thrust-to-weight ratio and maximum body rate, respectively. 

\subsubsection{Reward Function}

Analogous to the observation vector, the reward function in all three scenarios is also composed of two parts, defined as follows:
\begin{equation}
    {r} = {r}_{\text{general}} + {r}_{\text{specific}},
\end{equation}
where ${r}_{\text{general}}$ denotes the general reward component common to all three scenarios, while ${r}_{\text{specific}}$ is scenario-specific and will be introduced in section \ref{scenario customization}.

Specifically, the general reward component is designed as ${r}_\text{general} = {r}_\text{safe}+{r}_\text{crash}+ {r}_\text{smooth}$, where we define the sub-rewards as follows.

\begin{figure}[h]
    \centering
    \vspace{-1.2em}
    \subfigure[Visualization of $\phi$ and $\theta$, which are the payload's deviation projected onto the body frame's Y-Z plane and X-Z plane, respectively.]
    {\includegraphics[width=0.19\textwidth, trim = 0 85 650 0, clip]{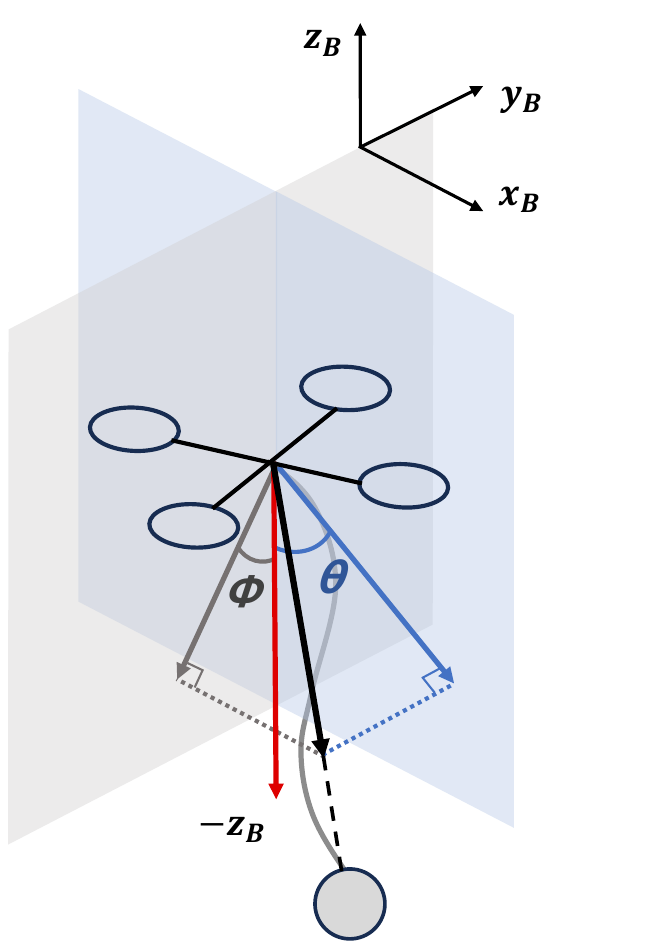}\label{cable-angle}}
    \hspace{0.1cm}
    \subfigure[Illustration of an unsafe state. When the payload swings to a high altitude, its large deviation angle exceeds the threshold ($|\theta| > \phi_{max}$), thereby incurring the ${r}_{\text{safe}}$ penalty.]{\includegraphics[width=0.26\textwidth, trim = 330 135 320 80, clip]{Figures/safe-reward.pdf}\label{unsafe example}}
    \vspace{-0.4em}
    \caption{Definition and formulation of the safety reward.}
    \vspace{-1.2em}
\end{figure}

The safety reward ${r}_{\text{safe}}$ is designed to mitigate the risk of catastrophic failure during agile flights. When the payload swings to a high altitude, the cable potentially entangles with the propellers. To this end, ${r}_{\text{safe}}$ imposes penalties when the payload's deviation angle $\phi$ or $\theta$ (Fig. \ref{cable-angle}) exceeds the threshold $\phi_{max}$ (Fig. \ref{unsafe example}), defined as:
\begin{equation}
    \begin{aligned}
        {r}_\text{safe}=
        \begin{cases}
            -r_{excess}, & |\phi| ~\text{or}~ |\theta| > \phi_{max}, \\
            0, & \text{otherwise}.
        \end{cases}
    \end{aligned}
\end{equation}

The crash reward ${r}_{\text{crash}}$ penalizes boundary violations with episodic termination strategies. Specifically, the crash reward activates exclusively when the quadrotor \textcolor{black}{position $\mathbf{x}_q$} or the payload \textcolor{black}{position $\mathbf{x}_p$} violates the spatially valid workspace $\mathcal{W}$, defined as:
\begin{equation}
    \begin{aligned}
        {r}_{\text{crash}}=
        \begin{cases}
            -r_{bound}, & \mathbf{x}_q ~\text{or}~ \mathbf{x}_p \notin \mathcal{W}, \\
            0, & \text{otherwise}.
        \end{cases}
    \end{aligned}
\end{equation}

The smoothness reward ${r}_{\text{smooth}}$ penalizes dynamically infeasible commands through limiting the abrupt variation of consecutive actions $\mathbf{a}$, defined as:
\begin{equation}
    {r}_{\text{smooth}} =  -\lambda_1 {\left \| \mathbf{a}_{t-1} - \mathbf{a}_{t} \right \|}.
\end{equation}

\subsection{Scenario customization}\label{scenario customization}
In this section, we detail the scenario-specific components of the observation vector and reward function for each of the three scenarios, thereby completing their respective RL problem formulations. The training pipeline and the descriptions of these scenarios are depicted in Fig. \ref{training pipeline}.

\subsubsection{Agile waypoint passing} \label{Agile waypoint-passing task}
The goal of this scenario is to establish a baseline for agile navigation, testing the policy's core capability for high-speed, safe traversal through a sequence of predefined waypoints.

To this end, the observation vector in the waypoint-passing scenario is defined as $ \mathbf{o}^{\textit{WP}}=[\mathbf{o}_{\textit{general}}, \mathbf{o}^{\textit{WP}}_{\textit{specific}}]^T$, where $\mathbf{o}^{\textit{WP}}_{\textit{specific}}=\mathbf{o}_w^{\textit{WP}}$ denotes the observation of waypoints. The waypoint observation is composed of the quadrotor's relative positions, $\textcolor{black}{\Delta \mathbf{x}_{q}^k}, k \in \{1,...,W\}$, to the next $W$ waypoints. We set $W=2$ to take two future waypoints into consideration, resulting in the waypoint observation vector $\mathbf{o}_w^{\textit{WP}} = [\textcolor{black}{\Delta \mathbf{x}_{q}^1, \Delta \mathbf{x}_{q}^2}]^\top \in \mathbb{R}^6$.

Furthermore, we design the reward function in this scenario as:
\begin{equation}
\begin{aligned}
    &{r}^{\textit{WP}} = {r}_{\text{general}} + \lambda_2r^{\textit{WP}}_{\text{specific}}, \\
    &{r}^{\textit{WP}}_{\text{specific}} = {r}^{\textit{WP}}_{\text{target}},
\end{aligned}
\end{equation}
where the target reward ${r}^{\textit{WP}}_{\text{target}}$ \textcolor{black}{at time step $t$} quantifies the spatial progress toward the next waypoint, incentivizing the system to reach the target in minimum time, a key characteristic of agile flight, defined as:
\begin{equation}
    r^{\textit{WP}}_{\text{target}} = \|\textcolor{black}{\Delta\mathbf{x}_{q_{t-1}}^1 }\|^2 - \|\textcolor{black}{\Delta\mathbf{x}_{q_{t}}^1}\|^2 . \label{equation-10}
\end{equation}

\subsubsection{Payload targeting}

In this scenario, we redefine the operational paradigm from direct quadrotor navigation to payload targeting. The primary objective shifts from guiding the drone itself to directing the payload to its destination, which requires actively leveraging the system's complex swing dynamics.

Similarly to scenario 1, the observation vector in payload-targeting scenario is defined as $ \mathbf{o}^{\textit{PT}}=[\mathbf{o}_{\textit{general}}, \mathbf{o}^{\textit{PT}}_{\textit{specific}}]^T$, where $\mathbf{o}^{\textit{PT}}_{\textit{specific}}=\mathbf{o}_w^{\textit{PT}}$ denotes the observation of waypoints for the payload. Meanwhile, we consider the observation of only one single future target in scenario 2. Accordingly, we define the waypoint observation as $\mathbf{o}_w^{\textit{PT}}=[\Delta \mathbf{x}_q,\Delta \mathbf{x}_p]^T$, where $\Delta \mathbf{x}_q$, $\Delta \mathbf{x}_p$ are defined as the quadrotor's and the payload's relative position to the target, respectively.

To manipulate the payload to reach a designated waypoint, we design the reward function as:
\begin{equation}
\begin{aligned}
    &{r}^{\textit{PT}} = {r}_{\text{general}} + \lambda_2r^{\textit{PT}}_{\text{specific}}, \\
    &{r}^{\textit{PT}}_{\text{specific}} = {r}^{\textit{PT}}_{\text{target}},
\end{aligned}
\end{equation}
where the target reward ${r}^{\textit{PT}}_{\text{target}}$ \textcolor{black}{at time step $t$} in this scenario quantifies the spatial progress of the payload, instead of the quadrotor, towards the designated waypoint, defined as:
\begin{equation}
    r^{\textit{PT}}_{\text{target}} = \|\Delta\mathbf{x}_{p_{t-1}} \|^2 - \|\Delta\mathbf{x}_{p_{t}}\|^2 .
\end{equation}

\subsubsection{Gate traversal}
This scenario assesses the policy's capability for highly aggressive maneuvers under tight spatial constraints, where collision-free passage requires exploiting the full system dynamics. In this scenario, we aim to achieve a collision-free navigation through a narrow gate, with a given final waypoint position $\mathbf{p}_{w}$ for the quadrotor and a known gate position $\mathbf{p}_g$.

The observation vector in the gate-traversal scenario is also defined as $ \mathbf{o}^{\textit{GT}}=[\mathbf{o}_{\textit{general}}, \mathbf{o}^{\textit{GT}}_{\textit{specific}}]^T$, where $\mathbf{o}^{\textit{GT}}_{\textit{specific}}=[\mathbf{o}_w^{\textit{GT}}, \mathbf{o}_g^{\textit{GT}}]^T$ encodes the observation of waypoints and the gate position, respectively. The waypoint observation $\mathbf{o}_w^{\textit{GT}}$ shares the identical \textcolor{black}{definition} with $\mathbf{o}_w^{\textit{WP}}$. Additionally, we define the gate observation $\mathbf{o}_g^{\textit{GT}}$ as the relative positions of the \textcolor{black}{quadrotor and payload} to the gate center $\mathbf{x}_{g}$, providing information about the gate's position.

To achieve collision-free navigation through a narrow gate, we design the reward function as:
\begin{equation}
\begin{aligned}
    &{r}^{\textit{GT}} = {r}_{\text{general}} + \lambda_2r^{\textit{GT}}_{\text{specific}}, \\
    &{r}^{\textit{GT}}_{\text{specific}} = {r}^{\textit{GT}}_{\text{target}} + \lambda_3{r}^{\textit{GT}}_{\text{gate}},
\end{aligned}
\end{equation}
where the target reward ${r}^{\textit{GT}}_{\text{target}}$ shares the identical definition with ${r}^{\textit{WP}}_{\text{target}}$ in Equation \ref{equation-10}, and the gate reward ${r}^{\textit{GT}}_{\text{gate}}$ incentivizes the drone to approach and traverse the gate with a two-stage design strategy, defined as:
\begin{equation}
    \begin{aligned}
        &r_i^{t,\text{gate}} = 
        \begin{cases}
            r_{arrival}, & \text{if pass the gate}, \\
            r_{guide}, & \text{otherwise}
        \end{cases} \\
        &r_{guide} = \frac{(\mathbf{p}_{w} - \mathbf{p}_g)}{\| \mathbf{p}_{w} - \mathbf{p}_g \| } \cdot \mathbf{v}_q,
    \end{aligned}
\end{equation}
where $r_{guide}$ incentivizes the magnitude of the drone's velocity projected onto the desired direction, which is defined as a vector from the gate position to the waypoint position.

\begin{figure*}[hb]
\vspace{-1.05em}
    \centering
        \subfigure[Suspended payload system motion in scenario 1 governed by two different policies. The policy trained with payload dynamics performs well, while the single-drone policy (trained without a suspended payload) fails to carry the payload in this scenario.]{\includegraphics[width=0.95\textwidth, trim = 120 355 120 0, clip]{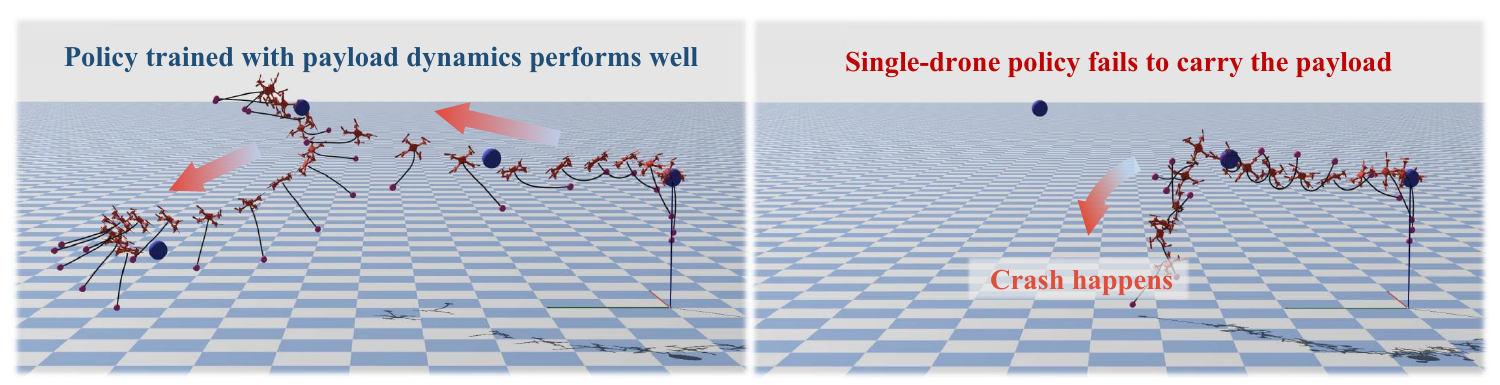}\label{sim_task1_a}} 
        \vspace{-0.2em}
        \subfigure[Visualized quadrotor and payload trajectories on different tracks (0.5m threshold). The results show that our method enables the suspended system to exhibit both excellent speed performance and remarkable agility.]{\includegraphics[width=0.93\textwidth, trim = 120 0 120 192, clip]{Figures/sim_task1.pdf}\label{sim_task1_b}}
    \vspace{-0.5em}
    \caption{Simulation results of scenario 1.}
    \label{sim_task1} 
\end{figure*}

\subsection{Policy training}

To ensure stable and efficient policy training in light of the varying scales of the input features, all observations are normalized prior to network processing: the quadrotor relative positions $\Delta\mathbf{x}_{q} \gets \Delta\mathbf{x}_{q} / \mathbf{k}_q$, the quadrotor velocity $\mathbf{v}_q \gets \mathbf{v}_q / \mathbf{k}_v$, the payload relative positions $\Delta\mathbf{x}_{p} \gets \Delta\mathbf{x}_{p} / \mathbf{k}_p$, the gate relative positions $\mathbf{o}_g \gets  \mathbf{o}_g/\mathbf{k}_g$ and the deviation angles of the payload $\mathbf{o}_{p} \gets \mathbf{o}_{p} / \mathbf{k}_{\phi}$, where the division is conducted element-wise, and $\mathbf{k}_q$, $\mathbf{k}_v$, $\mathbf{k}_p$, $\mathbf{k}_g$ and $\mathbf{k}_\phi$ are normalization constants listed in Table~\ref{tab:Simulation_Parameters}. 

\begin{table}[htbp]
    \caption{Simulation and Reward Parameters}
    \vspace{-0.9em}
    \begin{center}
        \renewcommand{\arraystretch}{1.2}
        \begin{tabular}{ll|ll}
            \Xhline{1pt}
            \textbf{Parameter} & \textbf{Value} & \textbf{Parameter} & \textbf{Value} \\ \hline
            $T_{\text{max}}$ & $3.5$ & $\boldsymbol{\omega}_{\text{max}}$ [rad/s] & $[15, 15, 5]$ \\
            $\mathbf{k}_q, \mathbf{k}_p$ [m] & $[5, 5, 1]$ & $\mathbf{k}_v$ [m/s] & $[10, 10, 3]$ \\ 
            $\mathbf{k}_{g}$ [m] & $[3, 3, 1]$ &  $\mathbf{k}_{\phi}$ [rad] & $[1.5, 1.5]$ \\ 
            $r_{{bound}}$ & $10$ & $r_{{excess}}$ & 3 \\ 
            $r_{{arrival}}$ & $20$ & $\lambda_1$ & $1 \times 10^{-4}$ \\ 
            $\lambda_2$ & $10$ & $\lambda_3$ & $5 \times 10^{-3}$ \\ \Xhline{1pt}
        \end{tabular}
    \end{center}
    \label{tab:Simulation_Parameters}
    \vspace{-1em}
\end{table}
With the overall framework illustrated in Fig. \ref{training pipeline}, our policy is trained using Proximal Policy Optimization (PPO) \cite{PPO}. The policy network, implemented as an MLP, consists of two hidden layers with 128 units each. A final projection layer then maps the resulting features to the normalized action space via a Tanh activation function. To better satisfy the Markov property, the network's input observation is augmented with the action from the previous timestep.

\textcolor{black}{Crucially, in all three scenarios, targets are uniformly resampled within a spatial volume of $[-2, 2] \times [-2, 2] \times [0.5, 1.5]$ m relative to the drone, triggered by either successful arrival or episode termination, to generate random sequences rather than fixed tracks. For Scenario 3, the gate is also spatially randomized and oriented primarily perpendicular to the path for feasibility, with random yaw perturbations ($\pm 15^\circ$) added to enhance robustness.}

The complex characteristics of the system typically incur a significant time cost. To address this, we leverage Genesis \cite{Genesis}, a state-of-the-art, GPU-vectorized simulator that provides high-performance quadrotor simulation. The cable is specifically simulated as a serial chain of interconnected rigid links, an approach similar to that in \cite{wang2024impact}. All simulations are executed with a timestep of 0.01 s. This setup allows the policies for Scenarios 1, 2, and 3 to converge after approximately 102.0, 122.9, and 251.9 million timesteps, which corresponds to only 0.45, 0.52, and 2.42 hours of training time, respectively.
Moreover, the system's complex dynamics may also introduce unmodeled factors that result in non-negligible sim-to-real gaps, such as unpredictable payload swings induced by external air disturbances and imperfect drone hovering. Therefore, we adopt a domain randomization strategy by incorporating random minor variations in the payload's initial deviation angle \textcolor{black}{(within $\pm 10^\circ$)}, to ensure the distribution of training data encompasses real-world conditions.

\section{SIMULATION RESULTS AND ANALYSIS}
This section presents the simulation results and further analysis of our approach. In scenario 1, we evaluate the performance of our trained policy across three tracks, utilizing a near-time-optimal policy for a single drone (without a suspended payload) on the same tracks as a reference. In scenario 2, we validate the method's ability to navigate the payload to a given target and demonstrate that the policy effectively exploits the cable's natural swing characteristics. Finally, in scenario 3, we benchmark the performance of our method against a state-of-the-art work \cite{wang2024impact} executing identical one-time gate traversals. All training and simulation experiments are performed on a workstation with an AMD Ryzen R9-9950 CPU and NVIDIA RTX 5090 D GPU. \textcolor{black}{For clarity, unless otherwise specified, all velocity metrics (Max Vel. and Avg Vel.) reported in the following tables (Table \ref{tab:task1}--\ref{tab:avg_performance}) refer to the magnitude of the quadrotor velocity.}
\subsection{Agile waypoint passing}

In this scenario, we validate the policy through numerous trials on tracks with varying complexity, where the drone is required to pass a given sequence of waypoints within a predefined radius.

\textit{\textcolor{black}{Agility analysis:}} We first rule out a naive approach: agile flights for the suspended payload system can be achieved simply by treating the payload as an external disturbance and relying on the inherent robustness of the RL method. To verify this, we additionally establish a reference policy only incorporating the drone dynamics, similar to the methods in \cite{song2021autonomous}, \cite{wang2025dashinggoldensnitchmultidrone}. When applied to scenario 1, the policy trained solely on the drone's dynamics is unable to complete the waypoint-passing task, often resulting in catastrophic failure (e.g. a crash), as illustrated in Fig. \ref{sim_task1_a}. In contrast, our policy, which accounts for the complete dynamics of the coupled system, successfully navigates all waypoints.

Furthermore, we conduct a focused analysis of the system's agility in this scenario. We navigate the suspended system through three tracks and demonstrate the trajectories in Fig. \ref{sim_task1_b}. We also provide the performance of the aforementioned reference policy when it operates a single drone (without a payload) on the same waypoint-passing tracks. The key metrics summarized in Table \ref{tab:task1} underscore the high agility of our method. While ensuring successful and safe task completion, our policy achieves an average maximum velocity of 88.38\% and an average velocity of 89.86\% relative to the reference policy, which is a near-time-optimal policy for single-drone navigation using the identical quadrotor configuration.

\begin{table}[htbp]
  \centering
  \vspace{-0.4em}
  \caption{Performance Metrics Across Different Simulation Tracks in Scenario 1 (0.5m Threshold)}
  \label{tab:task1}
  \begin{tabular}{@{\hspace{0.5em}}
                  c
                  c
                  S[table-format=1.2]
                  S[table-format=1.2]
                  S[table-format=2.2]
                  @{\hspace{0.5em}}}
    \toprule
    \multicolumn{2}{c}{\textbf{Configuration}} & \multicolumn{3}{c}{\textbf{Performance Metrics}} \\
    \cmidrule(lr){1-2} \cmidrule(lr){3-5} 
    \multirow{2}{*}{\textbf{Track}} & \textbf{Dynamic} & {\textbf{Max Vel.}} & {\textbf{Avg Vel.}} & {\textbf{Time}} \\
    & \textbf{Model} & {(m/s)} & {(m/s)} & {(s)} \\
    \midrule
    \multirow{2}{*}{Zigzag} & with payload  & 5.59 & 3.39 & 2.20  \\ [2pt]
                             & single drone & 6.47 & 3.75 & 1.68  \\
    \midrule
    \multirow{2}{*}{Eight - 3 laps} & with payload  & 5.64 & 3.38 & 11.96 \\ [2pt]
                             & single drone & 5.87 & 3.66 & 10.05 \\
    \midrule
    \multirow{2}{*}{Heart} & with payload  & 5.96 & 4.04 & 8.74  \\ [2pt]
                             & single drone & 7.11 & 4.62 & 7.42  \\
    \bottomrule
  \end{tabular}
  \vspace{-0.4em}
\end{table}

\begin{figure}[h]
    \centering
    \includegraphics[width=0.48\textwidth, trim = 0 60 0 25, clip]{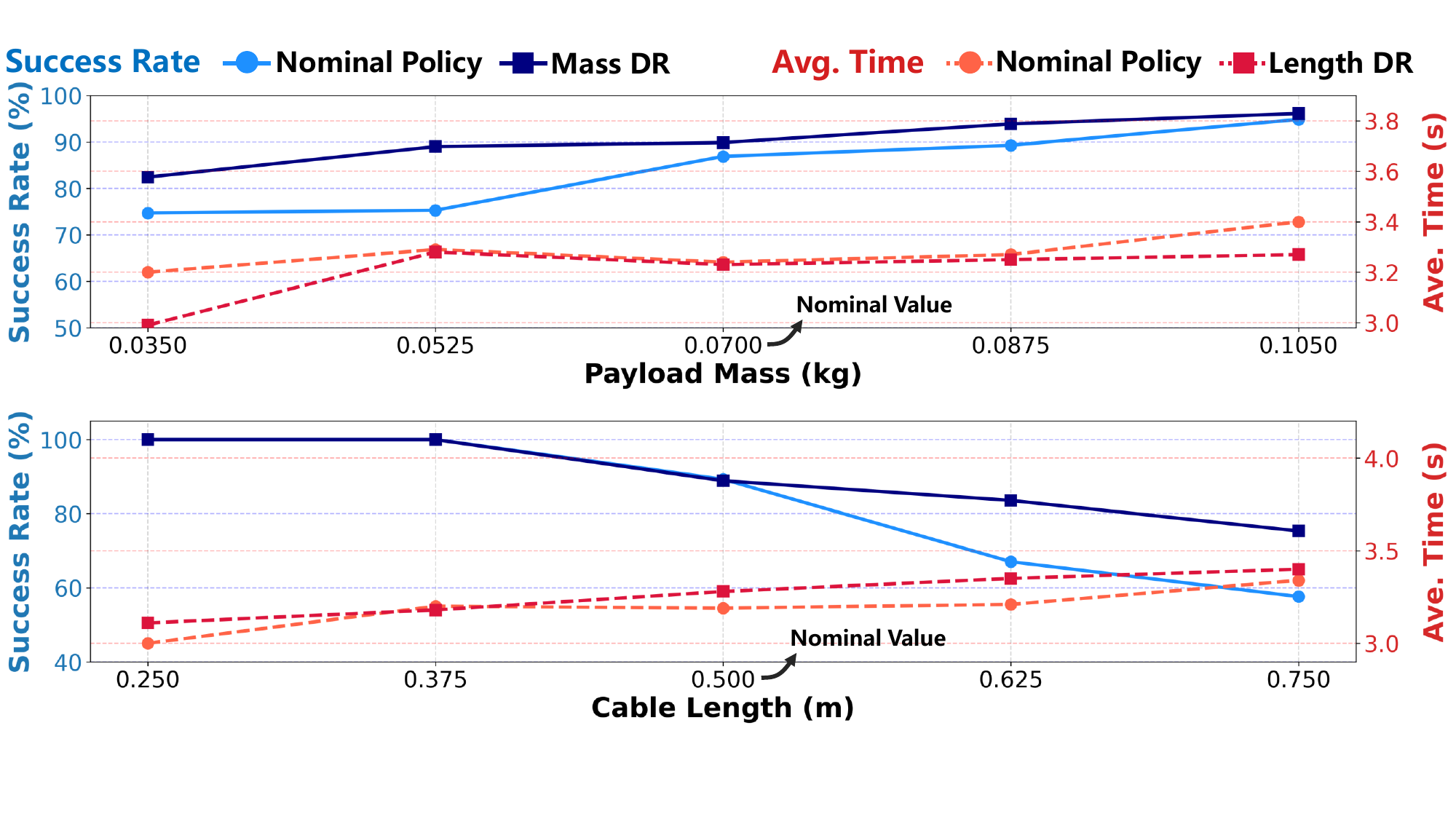}
    \caption{\textcolor{black}{  Robustness analysis regarding parameter variations in scenario 1. Success rates and average completion times are compared under varying payload masses (top) and cable lengths (bottom).}}
    \label{curve}
    \vspace{-0.9em} 
\end{figure}

\textit{\textcolor{black}{Robustness analysis:}} \textcolor{black}{We further evaluate adaptability to payload mass and cable length variations. Specifically, we compare the nominal policy against policies trained with extended domain randomization (extended DR) on these two parameters. Evaluations occur at uniformly selected test points spanning $\pm 50\%$ of nominal values, involving 100 random 8-waypoint tracks for each test point. Fig. \ref{curve} shows that success rates decline with lighter payloads or longer cables, attributed to reduced inertia or larger swing range, making the payload prone to swinging above the quadrotor and causing crashes. Moreover, while the nominal policy possesses inherent robustness, extended DR amplifies it, achieving a success rate above 80\% across all mass variations and boosting performance by approximately 20\% on the challenging long-cable conditions, with a slightly higher average completion time in the nominal condition.}

\subsection{Payload targeting}

In this scenario, we validate the effectiveness of navigating the payload to reach a given target.

\begin{figure}[h!]
    \centering
    \vspace{-1.2em}
    \subfigure[\textcolor{black}{The policy from scenario 1 yields a less compact trajectory}]
    {\includegraphics[width=0.2\textwidth, trim = 310 63 300 360, clip]{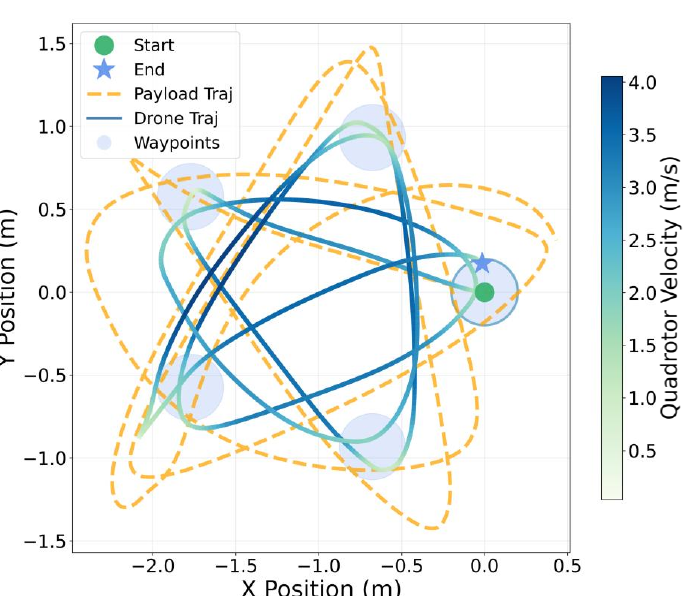}\label{genesis_task2}}
    \hspace{0.25cm}
    \subfigure[\textcolor{black}{The policy from scenario 2 exploits the swing characteristics of the cable.}]{\includegraphics[width=0.2\textwidth, trim = 310 380 300 30, clip]{Figures/sim_task2.pdf}\label{task2_method2}}
    \vspace{-0.55em}
    \caption{\textcolor{black}{Visualized quadrotor and payload trajectories (2 laps, 0.2m threshold) on the star-shaped track governed by different policies.}} \label{fig: scenario 2 traj comparison}
    \vspace{-0.9em}
\end{figure}

As shown in Fig. \ref{task2_method2}, our method successfully completes a sequence of 10 payload targets, forming a double-loop, star-shaped trajectory. Additionally, we compare the velocity performance and completion time with those of scenario 1, where the identical targets are treated as quadrotor waypoints, as summarized in Fig. \ref{genesis_task2} and Table \ref{tab:task2}.

\textcolor{black}{The trajectory comparison presented in Fig. \ref{fig: scenario 2 traj comparison} illustrates that the policy from scenario 2 yields a more compact trajectory. This compactness reveals that the policy guides the quadrotor to brake earlier, leveraging the cable's natural swing dynamics to reach the targets. This is further confirmed by Table \ref{tab:task2}, which demonstrates that the policy from scenario 2 completes the track with reduced total quadrotor flight distance, validating corresponding superior efficiency.}

\begin{table}[h!]
  \centering
  \vspace{-0em}
  \caption{Performance Metrics Comparison on 10 Identical Targets in Scenario 2 (0.2m Threshold)}
  \label{tab:task2}
  \begin{tabular}{c
                  S[table-format=1.2]
                  S[table-format=1.2]
                  S[table-format=1.2]
                  S[table-format=3.0]}
    \toprule
    \multirow{2}{*}{\textbf{Scenario}} & {\textbf{Max Vel.}} & {\textbf{Avg Vel.}} & {\textbf{Time}} & \textcolor{black}{\textbf{Quad. Dist.}} \\
     & {(m/s)} & {(m/s)} & {(s)} & \textcolor{black}{(m)}  \\
    \midrule
    1 & 4.05 & 2.55 & 8.57 & \textcolor{black}{21.85} \\ [2pt]
    2 & 3.33 & 1.78 & 8.15 & \textcolor{black}{15.72} \\
    \bottomrule
  \end{tabular}
  \vspace{-0.8em}
\end{table}

\subsection{Gate traversal}
\begin{table*}[htbp]
  \centering
  \caption{Performance Metrics Comparison Over 8 Simulation Trials in Scenario 3: Maximum Velocity, Traversal Time, Computational Time, Quadrotor Traversal Deviation, Payload Traversal Deviation, and Success Rate.}
  \label{tab:avg_performance}
  \begin{tabular}{c c c c c c c c}
    \toprule
    \multirow{2}{*}{\textbf{Method}} & \multicolumn{5}{c}{\textbf{Average Performance Metrics (Mean ± Std. Dev.)}} & \multirow{2}{*}{\textbf{Success Rate(\%)}} \\
    \cmidrule(lr){2-6}
     & {\textbf{Max Vel. (m/s)}} & {\textbf{Time (s)}} & {\textbf{Comp. Time (s)}} & {\textbf{Quad. Dev. (m)}} & {\textbf{Payload Dev. (m)}} & \\
    \midrule
    Impactor & 1.285 ± 0.087  & 2.082 ± 0.099 & 4.308 ± 1.418 & 0.181 ± 0.023 & 0.169 ± 0.055 & 100 \\ [4pt]
    Ours     & 3.673 ± 0.446 & 0.771 ± 0.082 & $\leq~$0.01 & 0.169 ± 0.041 & 0.144 ± 0.048 & 100\\
    \bottomrule
    \vspace{-2.4em}
    \label{sim_task3_tabular}
  \end{tabular}
\end{table*}

In this scenario, we evaluate the performance of our method with a challenging gate traversal maneuver. The objective is to guide both the quadrotor and suspended payload through a circular gate with a radius of 0.3 m, ensuring collision-free passage.
\begin{figure}[h]
    \centering
    \includegraphics[width=0.35\textwidth, trim = 0 185
     480 128, clip]{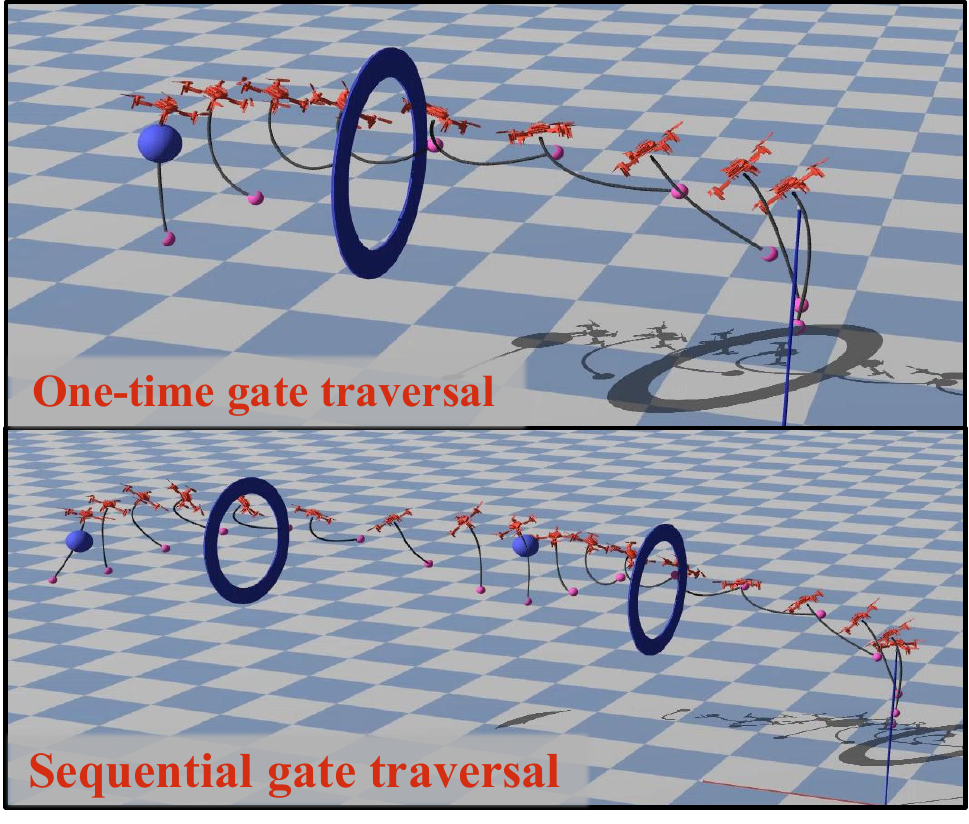}
    \vspace{-0.2em}
    \caption{System motions of gate traversal in scenario 3.}
    \label{sim_task3_a}
    \vspace{-1.0em} 
\end{figure}

\begin{figure}[h]
    \centering
    \vspace{-0.3em}
    \includegraphics[width=0.44\textwidth, trim = 415 165 10 120, clip]{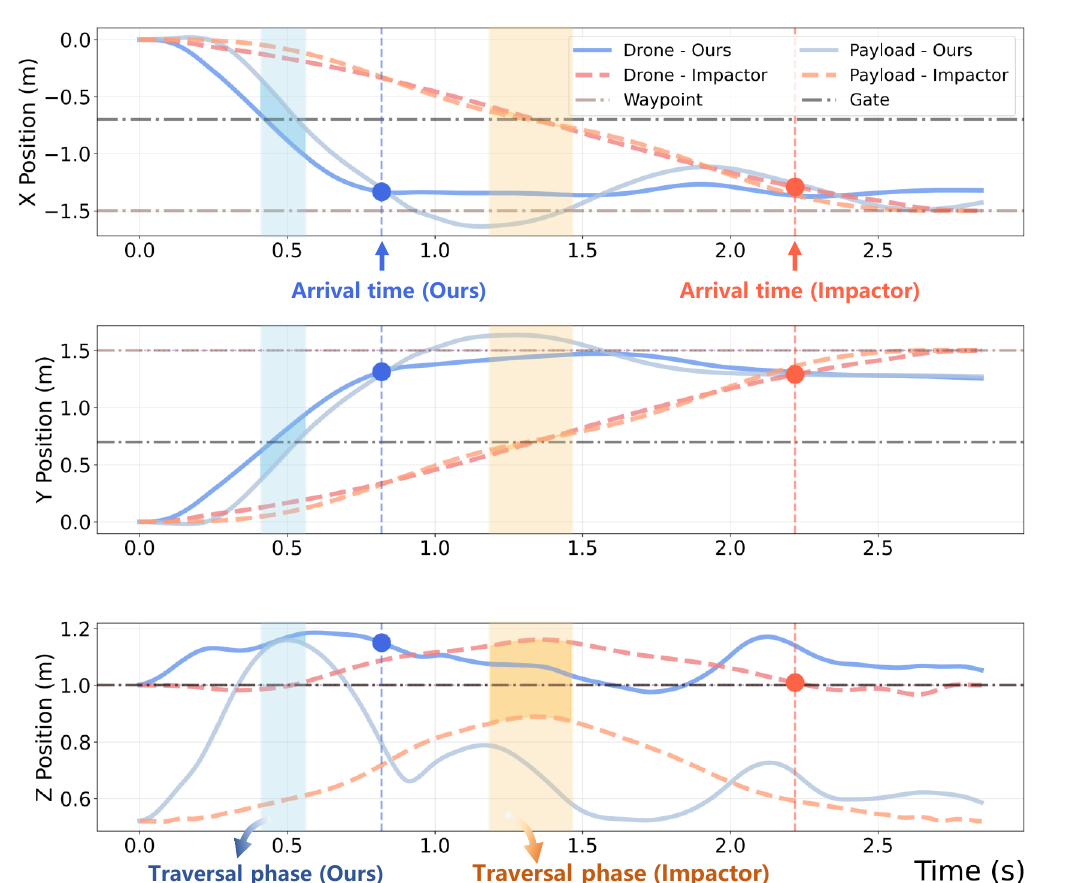}
    \caption{Time-series comparison of the trajectories for our method versus Impactor \cite{wang2024impact} \textcolor{black}{in scenario 3 (gate traversal)}. The shaded regions highlight the traversal phase (blue: our method; orange: Impactor), defined as the duration from the system's initial entry to its full exit of the gate.}
    \label{sim_task3_b}
    \vspace{-1.5em} 
\end{figure}

We validate our method's ability by performing both one-time and sequential gate traversal, as shown in Fig. \ref{sim_task3_a}. For performance evaluation, we benchmark our approach against \textcolor{black}{the aforementioned state-of-the-art optimization-based} Impactor \cite{wang2024impact} in the one-time traversal setting. \textcolor{black}{While effective for mode-switching planning, Impactor suffers from high computational costs, preventing its online deployment.} This comparison is conducted across 8 trials, where the final waypoint is positioned within the range of [-2.0, 2.0] m. We evaluate both methods based on 6 important metrics as presented in TABLE \ref{sim_task3_tabular}. To offer a more detailed view, Fig. \ref{sim_task3_b} illustrates the quadrotor and payload trajectories of a representative track governed by these two methods.

The experimental results indicate that both methods can successfully navigate all 8 trials without collision, and the deviations from the gate center are numerically similar. However, our method exhibits significantly greater agility, attaining a peak velocity 3x higher than Impactor and completing the trajectories in approximately 1/3 of the time. Notably, although Impactor can autonomously switch cable mode (slack or taut), this optimization is computationally demanding. Conversely, our deployed policy navigates the suspended payload system at 100 Hz (matching the training frequency, but flexible in practice), achieving real-time planning and control. This capability enables our approach to handle more complex and dynamic scenarios effectively.

\section{EXPERIMENT SETUP AND RESULTS}
\begin{figure}[hbt]
    \centering
    \vspace{-0.1em}
    \includegraphics[width=0.25\textwidth,trim = 0 0 420 0, clip]{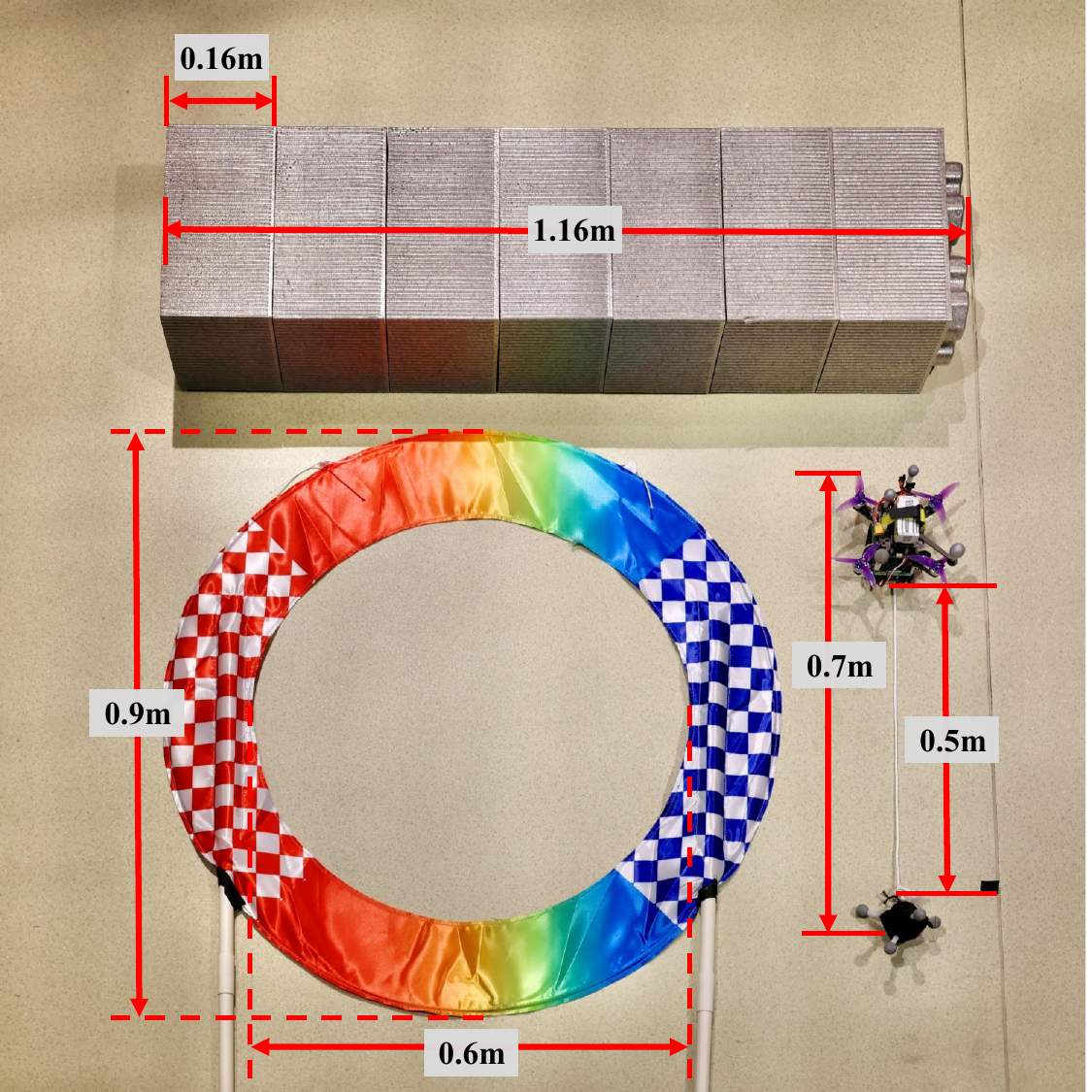}
    \vspace{0.2em}
    \caption{  Quadrotor suspended payload system and real-world setup, including a bar target (for scenario 2) and a circular gate (for scenario 3). Notably, the combined size of the system exceeds the gate's diameter, prohibiting a direct traversal and necessitating agile maneuvers.}
    \label{real-config}
    \vspace{-0.35em}
\end{figure}

For the real-world flights, we use a 305~g quadrotor (with a thrust-to-weight ratio of 3.5) equipped with a 70~g payload, as depicted in Fig. \ref{real-config}. The quadrotor runs Betaflight for low-level control and uses a Cool Pi computer for onboard policy inference at 100 Hz via LibTorch. An Optitrack motion capture system provides position and attitude measurements in a $5.0\,\text{m} \times 5.0\,\text{m} \times 2.0\,\text{m}$ space. Furthermore, we directly deploy the simulation-trained network without fine-tuning.
\subsubsection{Agile waypoint passing}
\begin{figure}[h]
    \centering
    \vspace{-0.5em}
    \includegraphics[width=0.36\textwidth,trim = 0 0 415 0, clip]{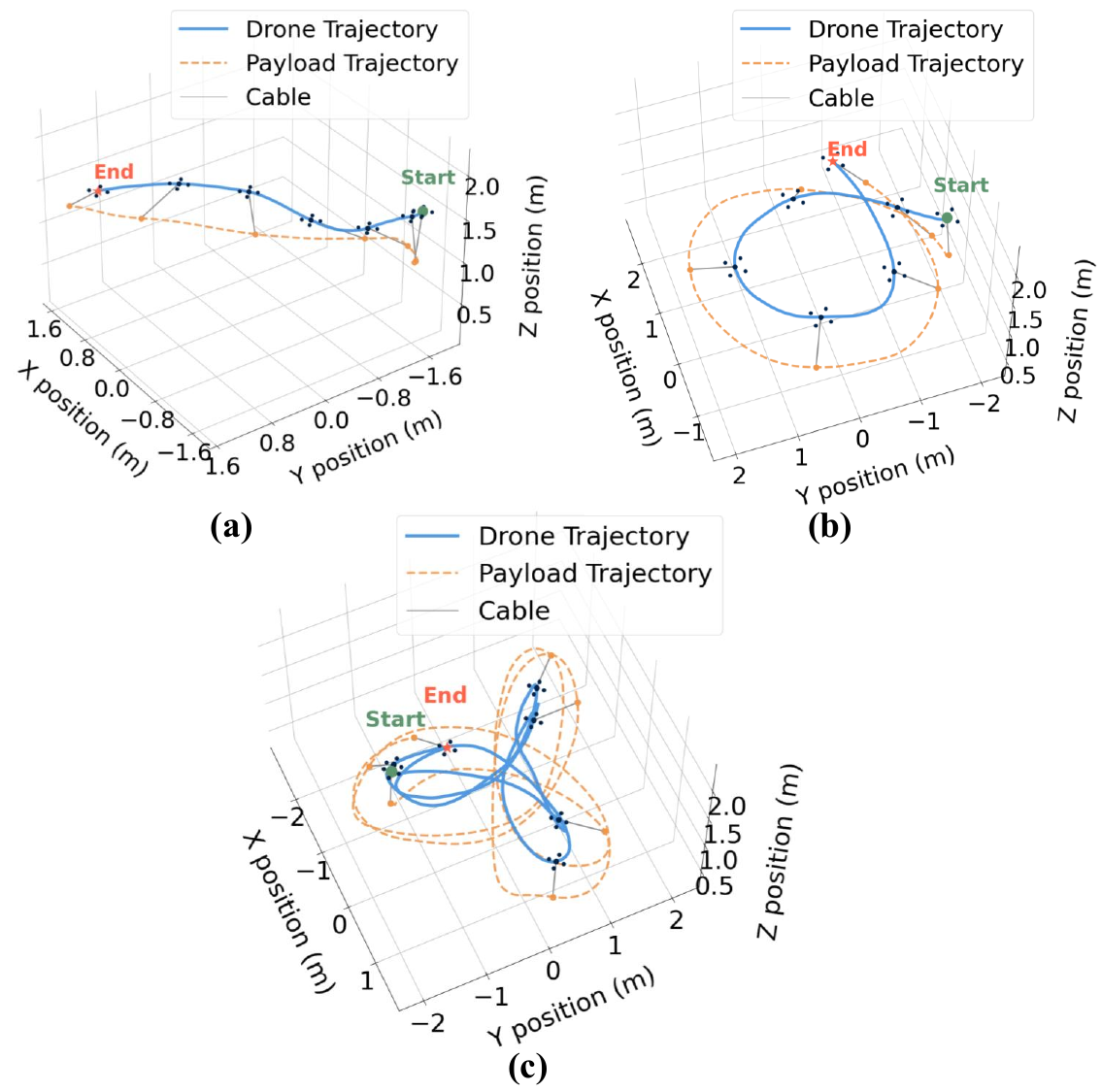}
    \vspace{-0.5em}
    \caption{  The 3D trajectories of both the quadrotor and the payload on 3 real-world tracks in scenario 1.}
    \label{real-task1-tracks}
    \vspace{-0.5em} 
\end{figure}

We design three distinct tracks to evaluate the performance of the learned policy in scenario 1, as one representative trajectory shown in Fig. \ref{real-exp-fig}(a). The 3D trajectories of both the quadrotor and the payload, presented in Fig. \ref{real-task1-tracks}, demonstrate that the policy can effectively tackle diverse tracks in the waypoint-passing scenario.

\begin{table}[h]
    \centering
    \caption{Performance Comparison of Simulation and Real-World Experiments on 3 Different Tracks \textcolor{black}{in Scenario 1.}}
    \label{tab:sim_vs_real}
    
    \begin{tabular}{cccc}
    \toprule 
    
    \textbf{Track} & \textbf{Experiment Type} & \textbf{Time (s)} & \textbf{Max Vel. (m/s)} \\
    \midrule 
    
    \multirow{2}{*}{\makecell{Track 1\\(4 waypoints)}} & Simulation & 1.33 & 4.71  \\
    \cmidrule(l){2-4} 
    & Real-World & 1.24 & 5.29  \\
    \midrule 
    
    \multirow{2}{*}{\makecell{Track 2\\(7 waypoints)}} & Simulation & 2.36 & 5.40 \\
    \cmidrule(l){2-4}
    & Real-World & 2.56 & 5.80  \\
    \midrule 
    
    \multirow{2}{*}{\makecell{Track 3\\(14 waypoints)}} & Simulation & 6.00 & 4.72  \\
    \cmidrule(l){2-4}
    & Real-World & 6.39 & 4.86  \\
    
    \bottomrule 
    \end{tabular}
    \vspace{-1em}
\end{table}

Additionally, to assess the sim-to-real transfer performance of the learned policy, we also compare the results of the aforementioned tracks as executed in both simulation and the real world. The quantitative results, summarized in Table \ref{tab:sim_vs_real}, confirm a successful policy transfer, showing that the real-world performance is highly comparable to the simulation with minor deviations.

\subsubsection{Payload targeting}
The performance of the learned policy in scenario 2 is assessed by its ability to reach a given target at a distance of 1.5 m, as shown in Fig. \ref{real-exp-fig}(b). The motion data recorded in several trials shows that the policy achieves an average payload targeting error of 0.12 m. The payload's average and maximum velocities are 1.61 m/s and 3.74 m/s, respectively. The quadrotor's average and maximum velocities are 1.42 m/s and 2.98 m/s, respectively. These results indicate that the system exhibits high targeting accuracy and excellent speed performance in this scenario.

\subsubsection{Gate traversal}
The system's gate traversal capability in scenario 3 is assessed by requiring it to navigate through both a single gate and two consecutive gates, as shown in Fig. \ref{real-exp-fig}(c). Notably, this scenario is particularly demanding as the combined size of the quadrotor and swinging payload exceeds the gate's diameter, necessitating agile maneuvers, as the specific configuration detailed in Fig. \ref{real-config}.

In this scenario, both the quadrotor and the payload successfully navigate through narrow gates. During the consecutive traversal, the deviations from the center of the first gate are 0.21 m (quadrotor) and 0.17 m (payload), followed by deviations of 0.25 m and 0.26 m at the second. The entire maneuver is completed in a swift 1.57 s, with the quadrotor achieving an average velocity of 2.19 m/s and a maximum velocity of 5.72 m/s. These results underscore our policy's remarkable agility and safety.

\section{CONCLUSIONS}
In this letter, we introduce FLARE, a reinforcement learning framework designed to solve the challenging problem of agile flight for the quadrotor suspended payload system. Our approach, trained entirely in simulation, is validated across three demanding scenarios, demonstrating successful zero-shot sim-to-real transfer. This results in superior agility and computational efficiency compared to traditional optimization-based approaches.

This work highlights that model-free RL is an effective and practical approach for unlocking the vast agility potential of the suspended payload system. Future work could extend this framework to handle dynamic obstacles or more complex interaction tasks.

\bibliographystyle{ieeetr}

\end{document}